\documentclass[letterpaper]{article} 
\usepackage{aaai24}
\usepackage{times}  
\usepackage{helvet}  
\usepackage{courier}  
\usepackage[hyphens]{url}  
\usepackage{graphicx} 
\usepackage{natbib}  
\usepackage{caption} 
\usepackage{amsmath}
\usepackage{booktabs}
\usepackage{multirow}
\usepackage{newfloat}
\usepackage{listings}
\DeclareCaptionStyle{ruled}{labelfont=normalfont,labelsep=colon,strut=off} 
\title{CTP-Net: Character Texture Perception Network for \\Document Image Forgery Localization}
\author{
    Xin Liao \textsuperscript{\rm 1},
    Siliang Chen \textsuperscript{\rm 1},
    Jiaxin Chen \textsuperscript{\rm 1}\thanks{Corresponding author: Jiaxin Chen (chenjiaxin@hnu.edu.cn).},
    Tianyi Wang \textsuperscript{\rm 2},
    Xiehua Li \textsuperscript{\rm 1},
}
\affiliations{
    \textsuperscript{\rm 1}The College of Computer Science and Electronic Engineering, Hunan University, Changsha 410082, China.\\

    \textsuperscript{\rm 2}The Department of Computer Science, The University of Hong Kong, China.
}

\usepackage{bibentry}

\begin{document}

\maketitle

\begin{abstract}
Due to the progression of information technology in recent years, document images have been widely disseminated on social networks. With the help of powerful image editing tools, document images are easily forged without leaving visible manipulation traces, which leads to severe issues if significant information is falsified for malicious use. Therefore, the research of document image forensics is worth further exploring. In this paper, we propose a \underline{C}haracter \underline{T}exture \underline{P}erception Network (CTP-Net) to localize the forged regions in document images. Specifically, considering the characters with semantics in a document image are highly vulnerable, capturing the forgery traces is the key to localize the forged regions. We design a Character Texture Stream (CTS) based on optical character recognition to capture features of text areas that are essential components of a document image. Meanwhile, texture features of the whole document image are exploited by an Image Texture Stream (ITS). Combining the features extracted from the CTS and the ITS, the CTP-Net can reveal more subtle forgery traces from document images. Moreover, to overcome the challenge caused by the lack of fake document images, we design a data generation strategy that is utilized to construct a Fake Chinese Trademark dataset (FCTM). Experimental results on different datasets demonstrate that the proposed CTP-Net is able to localize multi-scale forged areas in document images, and outperform the state-of-the-art forgery localization methods, even though post-processing operations are applied.
\end{abstract}

\section{Introduction}
\begin{figure}[!t]
	\centering
	\includegraphics[width=0.48\textwidth]{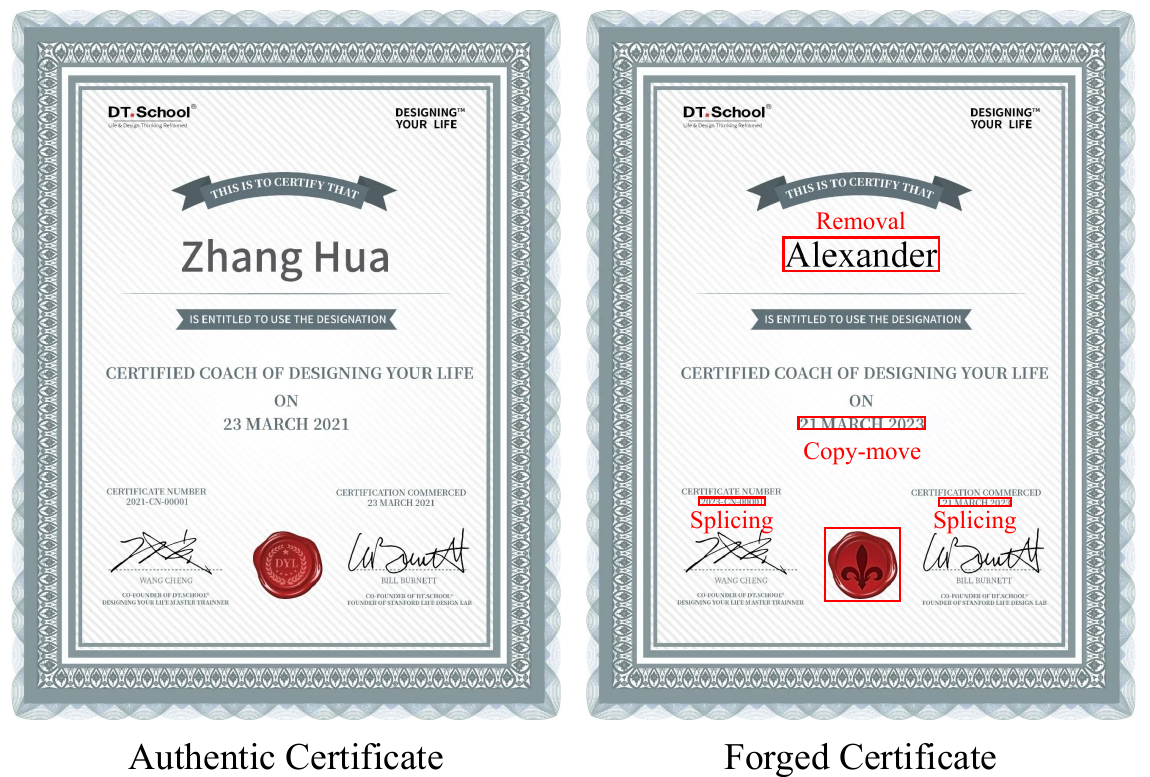}
	\caption{The example of a digital coaching certificate image (left) being forged (right) by using Photoshop. The name, date, and stamper of the certificate have been tampered with, which are boxed in red.}
	\label{fig_1}
\end{figure}

Digital images have been evolved into an ingenious foundation for information representation due to the explosion in the number of image acquisition devices and the wide use of social media. As a representative category of digital images, document images have been widely used in electronic identification, digital certificate, and cyber commerce, playing an essential role in human society \cite{Chen'DigitalDocUse}. Unfortunately, document images can be easily forged with the development of image processing applications, including Photoshop, CorelDRAW, and Fireworks. It is challenging to distinguish their authenticity from direct observation. Moreover, it has been demonstrated that characters in document images can be edited with end-to-end convolutional neural networks \cite{Zhao'ADI}. As shown in Fig. \ref{fig_1}, the pristine coaching certificate contains essential information such as name, date, and stamper. The information is forged by Photoshop and the tampering traces in the forged certificate are almost invisible to naked eyes. One can seek illegitimate interests with the fake certificate image. Furthermore, a slight change in a sentence might twist the entire semantic information. Therefore, localizing forgeries in document images has become an urgent issue.

Many natural image forgery detection and localization methods have been proposed over the past decades to distinguish between original and maliciously processed images \cite{Zhou'LearnignRichFeature,Li'SemanticChanged,Bappy'HLSTM,Chen'MVMSS,Wu'MantraNet,Cozzolina'Noiseprint,Chen'SNIS}.
Unlike natural images, characters are the main component of a document image, which consists of slender lines. It is a challenge to exploit character features from document images with complex background, negatively impacting tampering trace detection and resulting in poor document image forgery localization performance.

With the increasing use of document images, forensic technology for locating tampered regions in document images has attracted growing attention from researchers in recent years. The previous related works on document image forensics can be summarized in two categories: forgery detection methods \cite{Wu'PrinterForensics,Gupata'ForenssicDocSystem} and forgery localization methods \cite{Shang'DocForensic,Zhuang'DFCN,Xu'DocForensicTwoStream}. However, most studies learned the forgery traces by a manually designed characteristic operator, such as specific tampering traces based on geometric and noise distortion. The capability of those forensic methods is limited by the handcrafted characteristic operator. The forged regions in document images might be tiny if only one character is manipulated maliciously. Furthermore, instead of performing tampering operations in a naive fashion, a forger would apply some post-processing operations to envelop forgery traces. Thus, despite the increasing emergence of document image forgery localization approaches, exposing tampered regions in document images is a practical challenge. It is necessary to design a more powerful forgery localization method for forged document images.

Moreover, with the purpose of privacy-preserving, there are few document image datasets publicly available in reality. A document image dataset named SACP is provided in an adversarial attack competition \cite{SACP}. However, some tampered regions are not strictly labeled in the dataset. The lack of training samples makes it a challenge to train a well-performed forgery localization model since machine learning, especially deep learning, is data-hungry. Collecting large-scale realistic falsified document images and corresponding ground truths is time-consuming and laborious. Therefore, it is imperative to design an effective method for emulating tampering operations and generating sufficient fake document images for facilitate training.

In this paper, we propose a \underline{C}haracter \underline{T}exture \underline{P}erception Network (CTP-Net) to localize tampered regions in forged document images by Character Texture Stream (CTS) and Image Texture Stream (ITS). When manipulating document images using semantic-focused operations, sensitive information in text regions such as names, dates, and signatures is most likely to be forged. By focusing on the texture features of characters in the text areas, it becomes easier to extract tampering traces from document images, even when the background areas are complex.
Thus, we design the CTS based on optical character recognition. The CTS pays more attention to the texture features in character regions, contributing to exploiting the tampering traces around the text. Besides, background areas and icons are also the components of a document image. The ITS is designed to capture image texture representation, which is beneficial to consider the texture features of the whole image and improve the forgery localization performance. With the purpose of training models in the lack of realistic fake document images, we design a data generation strategy to imitate images falsified in reality and create large-scale tampering document images. The strategy makes it possible to collect sufficient training document images conveniently. The main contributions of this paper are as follows:
\begin{enumerate}
\item{We propose a novel two-stream network architecture, CTP-Net, to localize the forgery regions of document images. Instead of only extracting tampering features from the whole document image, we capture the texture features from both text regions and the entire document image. Inspired by text feature extraction in optical character recognition, CTP-Net down-samples the input images to capture the texture features at the character level.}
\item{An effective strategy to generate large-scale forged document images is provided to address the issue of a few public forgery document image datasets. The Fake Chinese Trademark (FCTM), which contains over 4,000 trademark images and corresponding ground truths, is constructed through this strategy.}
\item{Experimental results demonstrate that CTP-Net achieves promising performance in localizing tampered regions compared to the state-of-the-art methods even though the forgery regions are tiny. In addition, CTP-Net is robust against post-processing operations, such as resizing, cropping, and additive Gaussian noise.}
\end{enumerate}

\section{The Proposed Network Architecture}
\begin{figure*}[!htb]
	\centering
	\includegraphics[width=1\textwidth]{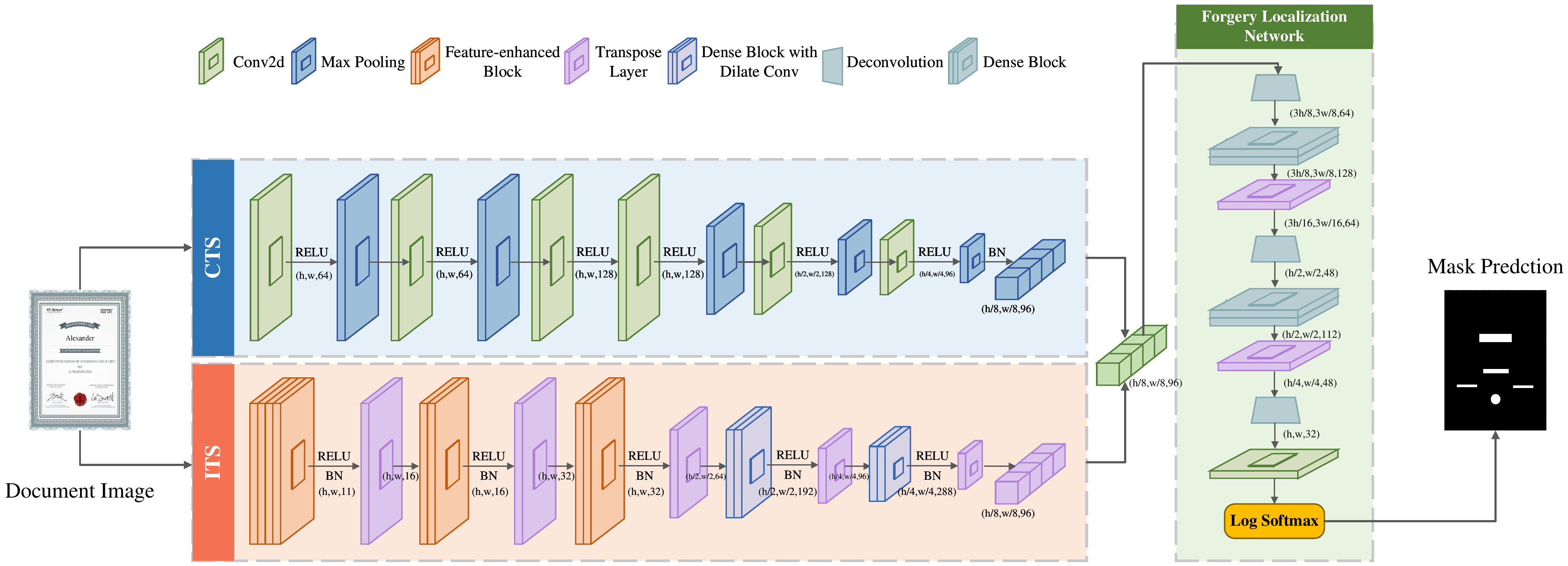}
	\caption{The proposed two-stream Character Texture Perception Net (CTP-Net) for document image forgery localization. The Character Texture Stream (CTS) is designed to capture imperceptible changes of malicious tampering manipulation in character texture information. The Image Texture Stream (ITS) is utilized to learn the forgery traces in image texture. The final mask prediction at the pixel level will be produced by the Forgery Localization Network.}
	\label{f0}
\end{figure*}

As shown in Fig. \ref{fig_1}, a document image contains text elements including name, date, and signature, and non-text elements such as backgrounds, and stampers. To improve the ability to perceive character, we need to consider the texture features of text and non-text elements independently, avoiding the mutual influence of text elements and non-text elements. As a result, the forgery traces can be better detected.

Fig. \ref{f0} illustrates our proposed document image forgery localization method. CTP-Net consists of three parts: the Character Texture Stream (CTS), the Image Texture Stream (ITS), and the Forgery Localization Network. It is noticed that capturing the feature of characters is the key to localize the forged regions since characters are the most easily forged element in document images. Therefore, the CTS is designed to capture the features of character texture based on optical character recognition, focusing on the tampering traces at the character level. As for the ITS, a novel feature-enhanced block can be regarded as a variant of the dense block \cite{Huang'DenseBlock}, which is designed to encourage feature reuse and strengthen feature propagation without increasing the cost of time and memory space. By stacking feature-enhanced blocks with regular convolution kernels and standard dense blocks with dilated convolution kernels, the ITS can capture the features of image textures, learning the tampering traces at the image level. Outputs of the two streams are fused and fed into the Forgery Localization Network, which consists of standard dense blocks and transpose layers. Finally, we obtain the mask prediction of document images and localize the tampered regions.

\subsection{Character Texture Stream}
Document images not only contain the characteristics of natural images but also include lots of character information. Besides, literal contents are more likely to be falsified when the document images are encountered with tampering attacks. Therefore, exploiting the features of literal contents is critical to document image forgery localization. Previous work \cite{Tang'ResearchOnText,He'FragNet,Shi'SceneTextRecognition} in text recognition that can be extended to document image forensic tasks shows that convolution neural networks are excellent in extracting character features in an image. Feature sequence extractors by taking the convolution and max-pooling layers from a standard CNN model are proposed to reveal the features of text \cite{Shi'SceneTextRecognition}. Those traditional character feature extractors used in text recognition focus on high-dimensional features through vast pooling operations to reveal the shape of characters including lines and contours. However, since it is usually supposed to down-sample the input image into high-dimensional features, the traditional character feature extractors may lose the information of inconsistency between the original and the forged characters, weakening the forgery detection performance. Instead of learning the representation of the characters, we aim to capture the tampering traces around the textures left by image manipulations.

To exploit this kind of anomaly feature, we skillfully design the Character Texture Stream (CTS). As shown in Fig. \ref{f0}, CTS is constructed by six convolution layers and five max-pooling layers. The kernel size of each convolution layer is 3$\times$3, followed by a max-pooling layer except the third layer, which is widely used in text feature extraction in optical character recognition. The RELU activation function is utilized to speed up the training. The deployment of convolution layers enhances the ability for nonlinear expression of the model. In addition, the stride of the first two max-pooling layers is set as 1, which amplifies the anomalous information without losing the resolution of the input. The last three max-pooling layers, whose stride is set as 2, can reduce redundant information about the feature, making the model focus on character texture. With convolution and pooling operations, the information unrelated to the character texture will be deserted. Therefore, the CTS can extract character texture features in document images even though the background is complex, which is the key to localizing the forgery regions in document images.

\subsection{Image Texture Stream}
Although text elements are the main components of a document image, non-text elements including icons and background textures still take risks of being falsified. Because non-text elements contain potential tampering traces, capturing texture features from the whole image is required. This subsection introduces the details of the Image Texture Stream (ITS) for revealing image texture features. With the development of deep learning in image forensics, dense blocks are beneficial for capturing subtle tampering traces in image texture \cite{Zhuang'DFCN}. However, the number of channels is increased linearly when concatenating the features in the original dense connection proposed in \cite{Huang'DenseBlock}. Since the number of parameters is increased, the cost of time and memory space is enlarged, making it challenging to learn image texture features. To overcome this problem, we change the connection mode between layers. Specifically, instead of stacking the features, the input of the next layer is produced by adding information to the output of the prior layer element-wisely, which keeps the dimension of the output feature the same as the input.

As shown in Fig. \ref{f0}, the architecture of ITS consists of three feature-enhanced blocks, two dilate dense blocks, and five transpose layers. The first three feature-enhanced blocks are modified densely connected convolution layers with standard convolution kernels. For a feature-enhanced block with $n$ internal convolutional layers, the output of the $i$-th layer is defined as
\begin{equation}
	\begin{aligned}
		f_{i}(h, w, c)=H_{i}(f_{i-1}(h, w, c)+f_{i-2}(h, w, c)),
	\end{aligned}
\end{equation}
where $f_{i}(h, w, c)$ represents the output feature of $layer_{i}$ and $(h, w, c)$ is the size of the feature map. $+$ represents the element-wise addition operation that combines the feature maps produced in $layer_{i-1}$ and $layer_{i-2}$. $H_{i}$ represents three sequential operations in the $i$-th layer: 3$\times$3 convolution, batch normalization, and RELU activation, and $i\in {1, 2, ...n}$.

Since the feature maps are fused by element-wise addition operations, the dimension of the output feature will be kept the same as the input, which makes full use of feature maps extracted by each layer without increasing the cost of training. The last two dilate dense blocks, which contribute to enriching the information in image texture features and obtaining better performance on forgery localization, are regular dense connection blocks with 3$\times$3 dilated convolution kernel, proposed in \cite{Yu'DilatedConv}. The dilation rate of each dilated convolution layer is set to be 2. Dilated convolution can help the model learn features from large regions to avoid volatile feature representations. To control the size of the feature map conducted by each block, four sequential operations are combined into a transpose layer which includes batch normalization, RELU activation, 1$\times$1 convolution, and pooling operation. Specifically, the pooling operation of the first two transpose layers is max pooling, whose stride is set to be 1. To reduce the number of parameters, three average pooling layers whose stride is set to be 2 are utilized in the last transpose layers.

\subsection{Forgery Localization Network}
After fusing the feature maps from the CTS and the ITS by element-wise addition operation, the forgery localization network is utilized to decode the combined feature map. As shown in Fig. \ref{f0}, the forgery localization network consists of three deconvolutional layers, two dense blocks, two transpose layers, a 1$\times$1 convolution layer, and the logic Softmax function. To obtain pixel-level predictions, the resolution of feature maps is supposed to be increased to the original input size gradually. It is proved that the deconvolutional layer \cite{Noh'Deconv} achieved better performance than the up sampling layer for precise object segmentation. Therefore, the employment of deconvolutional layers not only helps mask prediction to have the same shape as the image input but also contributes to localizing forgery regions in document images exactly. The final localization result can be obtained through a convolution layer with 1$\times$1 kernel size. Besides, each layer of the dense block consists of three sequential operations, 3$\times$3 convolution, batch normalization, and RELU activation. The size of each convolution layer is the same while the resolution of the feature map is increasing, which means that the forgery localization network concentrates on local anomalies gradually. Furthermore, the transpose layers used in the forgery localization network are the same as the last three in ITS, which contributes to reducing redundant information in the feature maps. A 1$\times$1 convolution layer is utilized to conduct a forgery localization prediction probability map of the input document image, which will be used to calculate the final mask prediction through the logic Softmax function.

To train the CTP-Net to achieve better performance on forgery localization, the cross entropy loss is adopted. When an iteration is completed, the forgery localization prediction probability map and the ground truth of the input document image will be used to calculate the loss. The target of the training is to minimize the loss.

\begin{figure*}[!htb]
	\centering
	\includegraphics[width=1\textwidth]{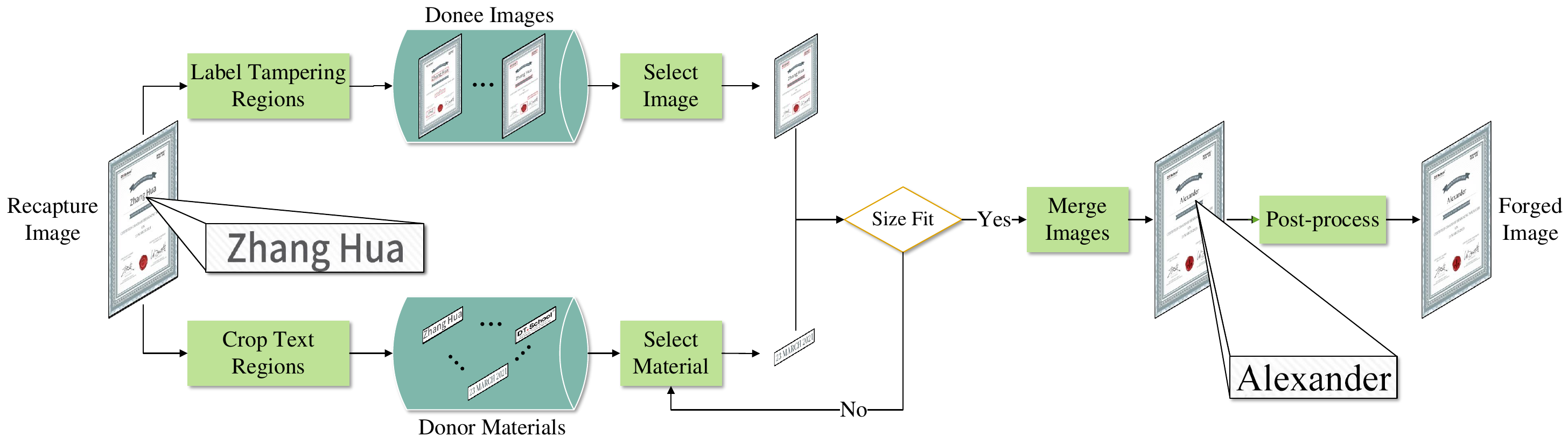}
	\caption{Diagram of Fake Chinese Trademark (FCTM) dataset generation strategy. First, the text information will be cropped from the recaptured images while the target regions are marked as duplicates of the originals. Then, the text information will be chosen to be merged into the donee image if it is matched with the target regions. Finally, Gaussian noise and JPEG compression are used to eliminate the forgery traces.}
	\label{operationFlow}
\end{figure*}

\section{Document Dataset Generation Strategy}
Since document images play an increasingly significant role in daily life, the research on document image forensics has attracted more attention. Zhuang et al. \cite{Zhuang'DFCN} built a book cover images dataset to simulate document images using Photoshop scripting. Xu et al. \cite{Xu'DocForensicTwoStream} conducted a document image dataset with multiple forgery methods. The Security AI Challenger Program (SACP) \cite{SACP} is an adversarial attack competition against document image forgery. A document image dataset named SACP is available in the competition.

Sufficient training samples are significant for a deep learning model to achieve excellent performance, and it takes work to collect large-scale training data. Unfortunately, only the SACP dataset is publicly available. The existing document-related dataset still has space for improvement. SACP only contains 3,005 document images, which is insufficient to train a well-performed model for document image forgery localization. Furthermore, the ground truths are generated by algorithms, leading to some mistakes in labeling tampered regions. To enlarge the number of training samples and improve the performance of the document image forgery detection and localization model, we build a fake document image dataset named Fake Chinese Trademark (FCTM).

The FCTM dataset consists of over 4,000 trademark images tampered with by Photoshop scripting and their corresponding ground truth. As the literary property of brands, the trademark is an intangible asset of a company or individual. The fake trademark is easily used for unlawful interests. The trademark images in FCTM are structured with white background areas and black characters, containing not only text elements, including the notary, address, and date, but also non-text elements, such as brands. These are typical characteristics of a document image. The resolutions of these trademark images are about 750$\times$820. To simulate forgery manipulations in real-world scenarios, copy-move, splicing, and post-processing which includes noise adding and JPEG compression, are applied. The trademark images are recaptured from the website of the Trademark Office of China National Intellectual Property Administration \cite{FTCM}.
As for the privacy concerns, since the information of privacy is hidden when the trademark images are published on the website, there is no privacy leak when the trademark images are used to conduct FCTM.
To improve the performance of CTP-Net on localizing forgery regions in document images, we pre-train the model on FCTM and retrain the pre-trained model on the corresponding database.

The details of the generating strategy used to conduct FCTM are illustrated in Fig. \ref{operationFlow}. Specifically, original trademark images are first recaptured from the website. Then, to develop a forged image, text information is cropped from the recaptured images, which are the donor materials. At the same time, the target regions are marked in duplicates of the recaptured images, which are regarded as donee images. Afterward, donor materials are randomly selected to be merged into the target regions of donee images if they are matched with the target regions filled with related donor materials. Finally, to make the tampered areas more challenging to detect, Gaussian noise and JPEG compression are utilized to eliminate the manipulation traces. By simulating the manipulation of tampering, the document images generated through this strategy contain tampering traces introduced by Photoshop, which is beneficial to learn the representation of manipulation traces for the forgery localization model. Hence, this strategy can generate sufficient tampering samples, which is the key to training a well-perform tampering localization model. The generated FCTM dataset is available at the link: https://github.com/FCTMdataset/FCTM.


\section{Experimental Evaluations}
\subsection{Experiment Setup}
\textbf{Datasets.} We evaluate CTP-Net on FCTM, SACP \cite{SACP} and NIST \cite{NIST} datasets. The FCTM dataset contains 50 authentic trademark images recaptured from the Internet and 4,138 forgery document images generated using our strategy. Ground truth masks are also provided along with the forgery images. The SACP dataset contains 3,005 falsified images with corresponding ground truths, which can be divided into nine categories, such as contracts, business certificates, trademark licenses, etc. Unknown post-processing methods are used against the detection of forensic techniques. The NIST dataset contains 564 tampered images, which are falsified by copy-move, splicing, or removal. Besides, the contained images are post-processed with unknown image editing methods.

\textbf{SOTA Methods.} We compare our method with four state-of-the-art methods: MT-Net \cite{Wu'MantraNet}, NoiPri \cite{Cozzolina'Noiseprint}, DFCN \cite{Zhuang'DFCN}, and TS-Net \cite{Xu'DocForensicTwoStream}. For fairness, we compare CTP-Net with DFCN retrained on our training dataset. For MT-Net and NoiPri, we adopt their officially released model. In addition, we only refer to the results in the published method of TS-Net.

\textbf{Training and Testing Protocols.} We split the training, validation, and testing set with the ratio of 8:1:1. As the resolutions of training data are different, we resized the resolutions of the training data to 512$\times$512. Since document image tampering localization is a pixel-level binary classification problem, we adopt the following commonly used performance metrics: the F1-score (F1), the Intersection over Union (IoU), the Matthews Correlation Coefficient (MCC), and the Area under Curve (AUC).

\textbf{Implementation Details.} The proposed model is implemented with the Pytorch deep learning framework. The SGD optimizer is adopted to train the model. The initial learning rate is set as 0.0001, while the batch size is set to 6 due to the memory space limit. Additionally, the model is tested with the validation set at the end of each 10 epochs whose value is set to 200.

\subsection{Ablation Study}
\textbf{Impacts of FCTM.} It is difficult to train a well-performed forgery localization model with only a few forged document images. Fortunately, our forged document image generation strategy alternatively provides a practical training approach. We investigate how the generated document image dataset (FCTM) can further improve the forgery localization performance of CTP-Net and DFCN in this experiment.

We utilize the tampering trademark images in FCTM to pre-train the models and then retrain them with the images in SACP. Besides, we train other models only on the SACP dataset without pre-trained on FCTM.
As shown in Table \ref{tab_documentImage}, the retrained models of CTP-Net and DFCN achieve better performance on the SACP dataset than the models without pre-training in terms of four performance metrics. It is observed that pre-training on FCTM helps models accomplish a significant improvement in F1-score, IoU, MCC, and AUC, respectively. The preliminary experiments show that retraining a model pre-trained with fake trademark images in FCTM improves the performance of forgery localization. To achieve the best performance, the remaining experiments in this paper follow this training strategy.

\begin{table}[t]
\centering
\small
\renewcommand{\tabcolsep}{1mm}
	\begin{tabular}{cccccc}
		\toprule
		\multirow{2}{*}[-3pt]{Datasets} &
		\multirow{2}{*}[-3pt]{Methods} &
		\multicolumn{4}{c}{SACP} \\
		\cmidrule[0.5pt](r){3-6}
		\multicolumn{1}{c}{}
		& & F1-score & IoU   & MCC    & AUC   \\
		\cmidrule[1pt](r){1-6}
		\multirow{3}{*}[6pt]{w/o FCTM}
		&DFCN (TIFS'21)        &0.11      &0.07  & 0.11    &0.53 \\
		&Ours  & 0.41  & 0.28     & 0.39  & 0.71  \\
		\cmidrule[0.5pt](r){1-6}
		\multirow{3}{*}[6pt]{w FCTM}
		&DFCN (TIFS'21) & \textbf{0.38}  & \textbf{0.27}   & \textbf{0.39} & \textbf{0.66} \\
		&Ours            & \textbf{0.45} & \textbf{0.31}    & \textbf{0.43}  & \textbf{0.72} \\
		\bottomrule
	\end{tabular}
\caption{The localization performance of models trained with and without FCTM.}
\label{tab_documentImage}
\end{table}

\begin{table}[t]
	\centering
    \small
	\begin{tabular}{ccccc}
		\toprule
		\multirow{2}{*}[-3pt]{Methods}&
		\multicolumn{4}{c}{SACP} \\
		\cmidrule[0.5pt](r){2-5}
		\multicolumn{1}{c}{}
		& F1-score & IoU   & MCC   & AUC   \\
		\cmidrule[1pt](r){1-5}
		Single CTS                & 0.20     & 0.13  &0.24   & 0.57   \\
		Single ITS                 & 0.35     & 0.23  &0.34   & 0.67   \\
		CTP-Net (CTS+ITS)            &\textbf{0.45}      & \textbf{0.31}  &\textbf{0.43}   &\textbf{0.72}    \\
		\bottomrule
	\end{tabular}
\caption{The localization performance of the proposed two-stream method (CTS+ITS), the single CTS method, and the single ITS method.}
\label{tab_Ablation_result}
\end{table}

\textbf{Impacts of CTS and ITS.} In this experiment, we perform an ablation study on SACP to validate the effectiveness of CTP-Net and the single-stream models. The experimental results are shown in Table \ref{tab_Ablation_result}. It can be observed that the single CTS and ITS models could have limited localization performance, but the proposed two-stream (CTS + ITS) model outperforms them.

Document images consist of text elements and non-text elements, which are different from natural images in characteristics of texture, color, and shape features. The single CTS model mainly focuses on the features in the text area, so it might ignore the tampering traces around non-text elements. As for the single ITS model, it is designed to consider the texture features of the entire document images. However, the non-text elements such as background areas, and icons, have a negative impact on capturing the feature of characters, and the single ITS model cannot perform well either. CTP-Net with CTS and ITS achieves better forgery localization performance than the other models. CTS contributes to making CTP-Net sensitive to subtle features of characters, weakening the influence of non-text elements. Furthermore, ITS captures the texture features of the whole image and ensures that CTP-Net does not miss the tampering traces around the non-text elements of the document image. Thus, combining CTS and ITS improves localization performance.

\subsection{Comparisons to State-of-the-Art Methods}
In this experiment, we compare the performance of our proposed CTP-Net with four state-of-the-art methods on document image forgery localization. As shown in Table \ref{tab_comparativeInDoc}, when localizing the forged regions in document images, our CTP-Net outperforms the comparative methods on SACP in terms of all four performance metrics. On the SACP dataset, the F1-score, IoU, MCC, and AUC achieved by CTP-Net are 0.45, 0.31, 0.43, and 0.72, respectively. The reason for this result is that CTP-Net pays more attention to the texture in character regions when capturing the texture features of the entire document image. This contributes to better performance in localizing falsified regions since characters are the main component of a document image.

\begin{table}[t]
	\centering
    \small
	\begin{tabular}{ccccc}
		\toprule
		\multirow{2}{*}[-3pt]{Methods} &
		\multicolumn{4}{c}{SACP}  \\
		\cmidrule[0.5pt](r){2-5}
		\multicolumn{1}{c}{}
		& F1-score & IOU   &MCC   & AUC \\
		\cmidrule[1pt](r){1-5}
		TS-Net (IJIS'22)             & 0.30  & 0.19   & 0.29      & -  \\
		MT-Net (CVPR'19)              & 0.07  & 0.04   &0.08      & 0.52  \\
		NoiPri (TIFS'20)      & 0.07  & 0.04   & 0.05     & 0.52  \\
		DFCN (TIFS'21)         &0.38      & 0.27  &0.39    &0.66   \\
		Ours                 &\textbf{0.45}      & \textbf{0.31}  &\textbf{0.43}    &\textbf{0.72}        \\
		\bottomrule		
	\end{tabular}
\caption{The comparisons of the localization performance to the state-of-the-art methods on the SACP dataset.}
\label{tab_comparativeInDoc}
\end{table}

\begin{table}[t]
\centering
\small
\renewcommand{\tabcolsep}{0.4mm}
	\begin{tabular}{cccccc}
		\toprule
		\multirow{2}{*}[-3pt]{Metrics} &
		\multirow{2}{*}[-3pt]{Methods} &
		\multicolumn{4}{c}{SACP} \\
		\cmidrule[0.5pt](r){3-6}
		\multicolumn{1}{c}{}
		&  &categ 1 & categ 2   & categ 3   & categ 4 \\
		\cmidrule[1pt](r){1-6}
		
		\multirow{5}{*}[6pt]{F1-score}
		&MT-Net (CVPR'19)          &0.05  &0.09    &0.07   &0.11   \\
		&NoiPri (TIFS'20) &0.04  &0.06    &0.11   &0.10   \\
		&DFCN (TIFS'21)     &0.26  &0.40    &0.44   &0.58   \\
		&Ours             &\textbf{0.32}  &\textbf{0.44}    &\textbf{0.55}   &\textbf{0.62}   \\
		\cmidrule[0.5pt](r){1-6}
		\multirow{5}{*}[6pt]{IoU}
		&MT-Net (CVPR'19)          &0.03  &0.05    &0.04   &0.07   \\
		&NoiPri (TIFS'20) &0.03  &0.03    &0.07   &0.06   \\
		&DFCN (TIFS'21)     &0.18  &0.28    &0.31   &0.43   \\
		&Ours             &\textbf{0.22}  &\textbf{0.30}    &\textbf{0.39}   &\textbf{0.49}   \\
		\cmidrule[0.5pt](r){1-6}
		\multirow{5}{*}[6pt]{MCC}
		&MT-Net (CVPR'19)         &0.06  &0.10    &0.07   &0.10   \\
		&NoiPri (TIFS'20) &0.04  &0.05    &0.07   &0.07   \\
		&DFCN (TIFS'21)     &0.27  &0.41    &0.44   &0.47   \\
		&Ours             &\textbf{0.32}  &\textbf{0.43}    &\textbf{0.51}   &\textbf{0.55}   \\
		\cmidrule[0.5pt](r){1-6}
		\multirow{5}{*}[6pt]{AUC}
		&MT-Net (CVPR'19)        &0.52  &0.53    &0.52   &0.53    \\
		&NoiPri (TIFS'20) &0.52  &0.52    &0.53   &0.52    \\
		&DFCN (TIFS'21)    &0.61  &0.66    &0.67   &0.72    \\
		&Ours             &\textbf{0.68}  &\textbf{0.71}    &\textbf{0.75}   &\textbf{0.76}    \\
		\bottomrule
	\end{tabular}
\caption{The comparison of the localization models tested on different scales of tampered regions in document images.}
\label{tab_MutilscaleOfDoc}
\end{table}

\subsection{Performance for Multi-scale Tampered Regions}
In practice, the scales of forged regions in document images are uncertain, which is very important for forgery detection and localization. In this experiment, we validate the effectiveness of different scales of fake regions. Firstly, we calculate the area of tampered regions in a forged image by pixel, and the proportion of fake areas in the image is obtained. Furthermore, the $\textrm{ratio}_{\textrm{fake}}$ areas is given by
\begin{equation}
	\begin{aligned}
		\textrm{ratio}_{\textrm{fake}} =\frac{\textrm{Count}(\textrm{fake}(x,y))}{\textrm{Count}(\textrm{original}(x,y))},
	\end{aligned}
\end{equation}
where $\textrm{fake}(x,y)$ represents the forged pixel and $\textrm{original}(x,y)$ represents the real one.
Then, the testing data is divided into four categories according to the ratio of forged regions in images. As illustrated in the Appendix, the ratios of the forged regions in four categories are from 0$\%$ to 10$\%$ (categ 1), 10$\%$ to 20$\%$ (categ 2), 20$\%$ to 50$\%$ (categ 3), and 50$\%$ to 100$\%$ (categ 4), respectively. It is noticed that the number of each class is different. The proportion of the four types is 38$\%$, 35$\%$, 24$\%$, and 3$\%$, respectively.


As shown in Table \ref{tab_MutilscaleOfDoc}, our CTP-Net achieves the best performance in localizing the tampered regions in each kind of document image in terms of all metrics. Since the tampering operations are mainly focused on the characters in document images, the forged regions of document images consist of characters basically, regardless of the proportion of forgery areas. With the skillful appliance of down-sampling and convolution operations, CTP-Net can concentrate on the texture features in character regions and reveal forgery traces around the text. This might be the reason why CTP-Net has outstanding performance in detecting and localizing all kinds of scales of tampered regions. As shown in the results of the first and the fourth categories, the values of the four metrics achieved in the first category are lower than those in the fourth category. This is because other elements in document images affect the learning of the features of small forged targets. Therefore, it is worth to be noticed that detecting and localizing small-scale tampering regions from the whole image remain a challenging issue.

\subsection{Qualitative Comparison}
\begin{figure}[tb]
	\centering
	\includegraphics[width=0.41\textwidth]{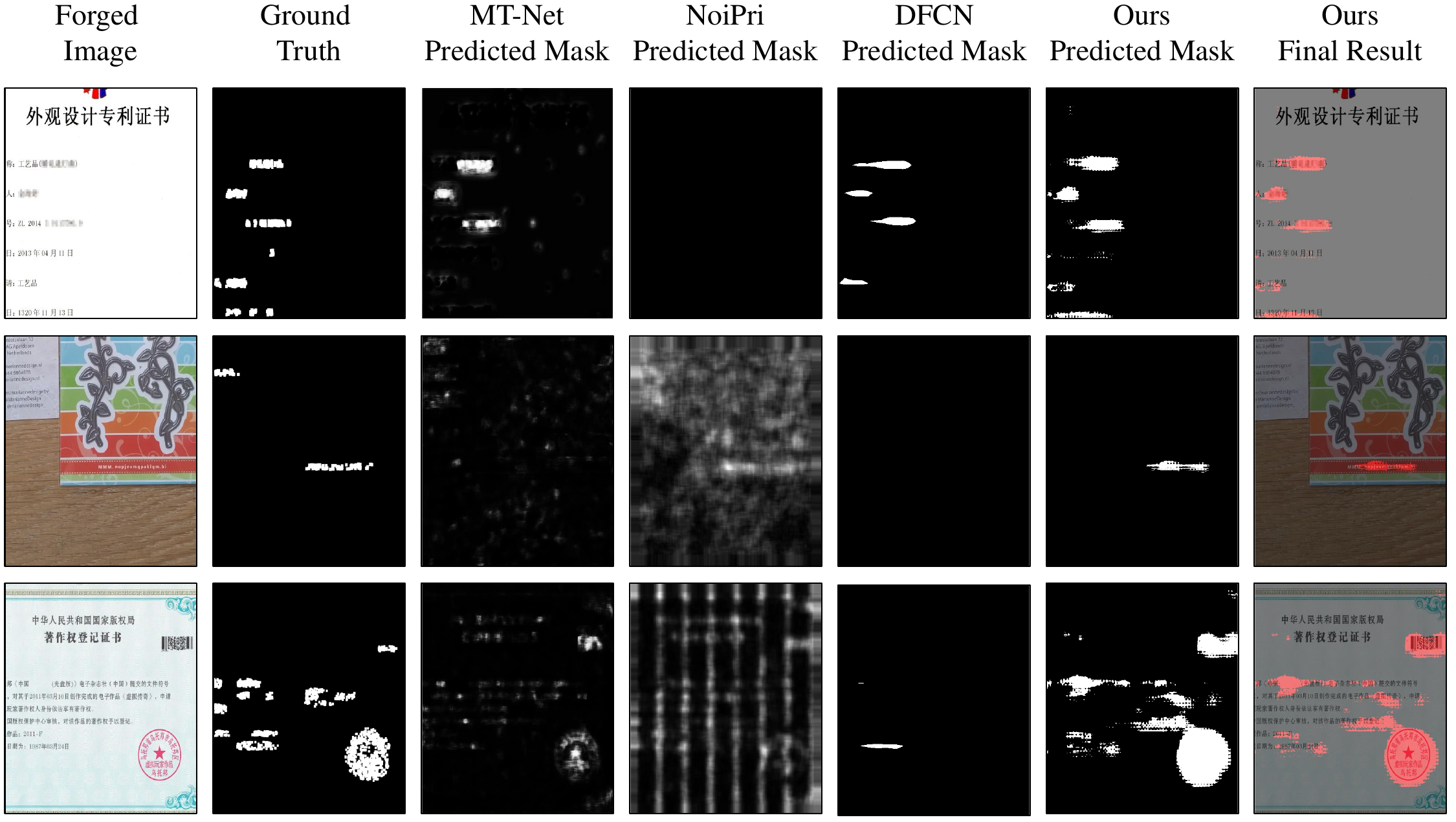}
	\caption{Qualitative results for document image forgery localization on SACP dataset.}
	\label{fig_resultsOfDoc}
\end{figure}

In Fig. \ref{fig_resultsOfDoc}, we provide some qualitative results of our proposed CTP-Net and SOTA methods in document image forgery localization. The images are selected from the SACP dataset. It can be observed that our CTP-Net is effective in obtaining better localization results than other methods. In detail, it can be found in the first two rows that the localization performance of the comparative methods on fake document images with complex background textures is lower than that on images with simple backgrounds, while our CTP-Net can achieve good performance in both cases. This is because the complex background has rich information, which is interference information when locating tampered regions and can affect the localization. CTP-Net extracts the texture features from character regions and the entire document image independently, which can weaken or even eliminate the interference effect of the other non-text elements. Thus, CTP-Net can expose the tampered regions from the complex background, obtaining promising forensic results.

\subsection{Robustness Analysis}
In this experiment, we evaluate the robustness of our CTP-Net by considering several common types of post-processing. We manipulate the testing data in the SACP dataset by applying resizing, cropping, or Gaussian noise with different factors. The AUC scores regarding different post-processing operations are shown in the Appendix. The performance of the detection methods is degraded when the resizing factor is lower than 1.0. It may be due to the fact that down-scaling has removed many image details which are important for forgery localization. Overall, it is observed that CTP-Net outperforms the SOTA methods in resisting resizing, cropping, and Gaussian noise.

\subsection{Performance for Natural Image Forgery Detection}
\begin{figure}[tb]
	\centering
	\includegraphics[width=0.31\textwidth]{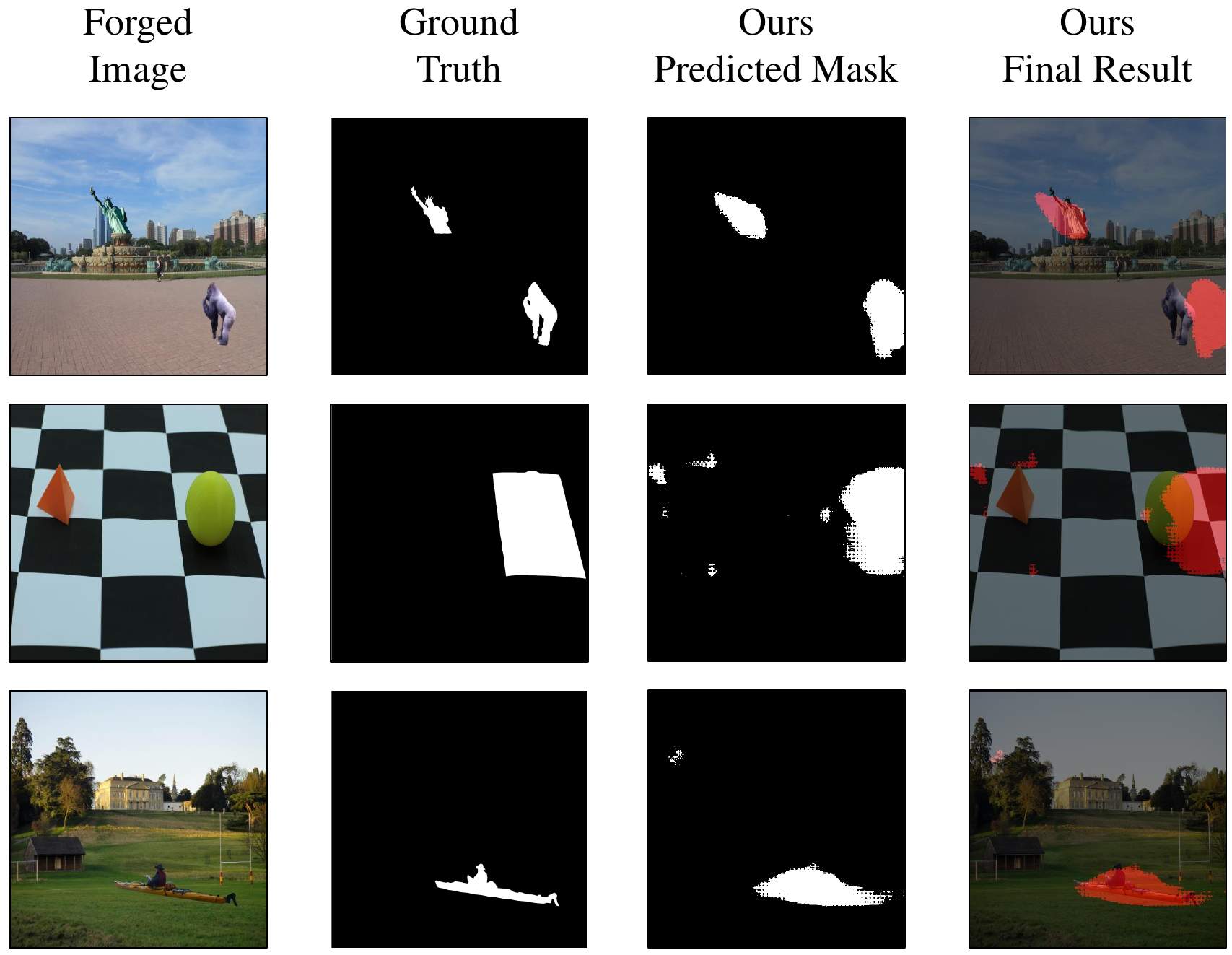}%
	\caption{The forgery localization performance for natural images on NIST dataset.}
	\label{fig_resultOfNist}
\end{figure}

In this experiment, we also conducted experiments on the natural images, which are selected from the NIST dataset. We retrain our CTP-Net on the NIST dataset based on the models pre-trained on the FCTM dataset. The F1-score, IoU, MCC, and AUC achieved by our CTP-Net are 0.33, 0.24, 0.30, and 0.70, respectively. Furthermore, in Fig. \ref{fig_resultOfNist}, we also provide some qualitative results for the evaluation of CTP-Net. It can be observed that CTP-Net can not only accurately localize forged document images, but also effectively detect forged natural images.

\section{Conclusion}
In this paper, we focus on tampering operations in document images and propose a method for document image forgery localization named CTP-Net, considering the fact that sensitive and essential information is easy to be forged. Since characters are the main object of document images, CTP-Net introduces the concept of optical character recognition to capture character texture further. Meanwhile, CTP-Net achieves better performance in localizing fraudulent regions by combining the texture features learned from character regions and the whole image. To deal with the lack of training data, we design a data generation strategy that has been adopted to build up the Fake Chinese Trademark (FCTM) dataset. Extensive experiments show that CTP-Net achieves state-of-the-art forged document image localization performance on multiple scales tampered regions. Moreover, CTP-Net has promising robustness against post-processing operations and can locate tampered regions in natural images.

\bibliography{aaai24}

\end{document}